\pdfoutput=1
\documentclass[11pt]{article}

\usepackage{acl}

\usepackage{times}
\usepackage{latexsym}

\usepackage[T1]{fontenc}
\usepackage[utf8]{inputenc}

\usepackage{microtype}

\newcounter{notecounter}
\newcommand{\enotesoff}{\long\gdef\enote##1##2{}}
\newcommand{\enoteson}{\long\gdef\enote##1##2{{
\stepcounter{notecounter}
{\large\bf \hspace{1cm}\arabic{notecounter} $<<<$ ##1: ##2 $>>>$\hspace{1cm}}}}}
\enoteson
\enotesoff % uncomment this to turn off the indexed notes

\usepackage{booktabs}
\usepackage{subcaption}
\usepackage{graphicx}
\usepackage{amssymb}
\usepackage{amsmath}
\usepackage{relsize}

\usepackage{booktabs}
\usepackage{multirow}
\usepackage{float}
\usepackage{xcolor}
\usepackage{rotating}
\usepackage{multirow}
\usepackage{pgfplots}
\usepackage{tikz}
\usepackage{xspace}

\usepackage{mathabx}
\definecolor{cmzhao}{rgb}{0.1, 0.8, 0.1}
\usepackage{enumitem}
\usepackage{graphicx}
\usepackage{booktabs}
\usepackage{multirow}
\usepackage{longtable}
\usepackage{amsmath}
\usepackage{cleveref}
\usepackage{url}
\usepackage{fancyvrb}

\newcommand{\pck}{PeaCoK\xspace}
\newcommand{\cmf}{ComFact\xspace}
\newcommand{\pxt}{PersonaExt\xspace}
\newcommand{\pep}{PersonaExt-PeaCoK\xspace}
\newcommand{\brak}[1]{{[}#1{]}}
\newcommand{\fullcol}[1]{\begin{minipage}{\textwidth}#1\vspace{0.5em}\end{minipage}}
\newcommand{\resizecol}[2]{\begin{minipage}{#1\textwidth}#2\vspace{0.5em}\end{minipage}}
\newcommand{\fitcol}[1]{\begin{minipage}{0.8\textwidth}#1\vspace{0.5em}\end{minipage}}
\newcommand{\fitcolm}[1]{\begin{minipage}{0.6\textwidth}#1\vspace{0.5em}\end{minipage}}

\title{Using Natural Language Inference to Improve Persona Extraction from Dialogue in a New Domain}

\author{
Alexandra DeLucia\textsuperscript{$1,2$}\thanks{\hspace{0.2cm}Work performed while interning at Sony.}, 
Mengjie Zhao\textsuperscript{$1$}, 
Yoshinori Maeda\textsuperscript{$1$},\\
{\bf Makoto Yoda}\textsuperscript{$1$},
{\bf Keiichi Yamada}\textsuperscript{$1$},
{\bf Hiromi Wakaki}\textsuperscript{$1$}\\
  \textsuperscript{$1$}Sony Group Corporation, Tokyo, Japan \\
  \textsuperscript{$2$}Center for Language and Speech Processing, Johns Hopkins University \\
  \texttt{aadelucia@jhu.edu} \\
  \texttt{\{Mengjie.Zhao, Yoshinori.B.Maeda, Makoto.Yoda,}\\
  \texttt{Keiichi.K.Yamada, Hiromi.Wakaki\}@sony.com} \\
}

\begin{document}
\maketitle

\begin{abstract}
While valuable datasets such as PersonaChat provide a foundation for training persona-grounded dialogue agents \citep{zhang-etal-2018-personalizing}, they lack diversity in conversational and narrative settings, primarily existing in the ``real'' world.
To develop dialogue agents with unique personas, models are trained to converse given a specific persona, but hand-crafting these persona can be time-consuming, thus methods exist to automatically extract persona information from existing character-specific dialogue \citep{wang-etal-2022-extracting}.
However, these persona-extraction models are also trained on datasets derived from PersonaChat and struggle to provide high-quality persona information from conversational settings that do not take place in the real world, such as the fantasy-focused dataset, LIGHT \citep{urbanek_learning_2019}.
Creating new data to train models on a specific setting is human-intensive, thus prohibitively expensive.
To address both these issues, we introduce a natural language inference method for post-hoc adapting a trained persona extraction model to a new setting.
We draw inspiration from the literature of dialog natural language inference (NLI; \citep{welleck-etal-2019-dialogue}), and devise NLI-reranking methods to extract structured persona information from dialogue. 
Compared to existing persona extraction models, our method returns higher-quality extracted persona and requires less human annotation.
\end{abstract}

%%%%%%%%%%%%%%%%%%%%%%%%
% Intro
%%%%%%%%%%%%%%%%%%%%%%%%
\section{Introduction}
\label{sec:intro}
\enote{mzhao}{We've finished the internal review, and there are two major comments:
1. why do we want to adapt to Peacok (relations), in distant narratives
2. what if there is no free-form text like utterances of a character, in some narratives.
I think Q1 is similar to enote-1 in the draft -- we could update the introduction to better motivate Peacok.
For Q2, I haven't realize this concern before, but I think we could mention this "cold start" problem in the Limitation section.}
\enote{mzhao}{In general, I think we have a good draft for submission; but I feel we may need more stress/discussions on commonsense persona -- e.g, do we want to adapt/specialize to \pck? You've mentioned that it has general relations, but I think it would be better if we make it more explicit}
Dialog agents are assigned a persona or a description of a character identity to impersonate while responding to a user.
These hand-crafted persona descriptions are supplied to the model's context along with the conversation history \citep{urbanek-etal-2019-learning,zhang-etal-2018-personalizing} or via character-specific embeddings added to the encoder \citep{li-etal-2016-persona} during response generation.

\begin{figure}[t]
    \centering
    \includegraphics[width=\columnwidth]{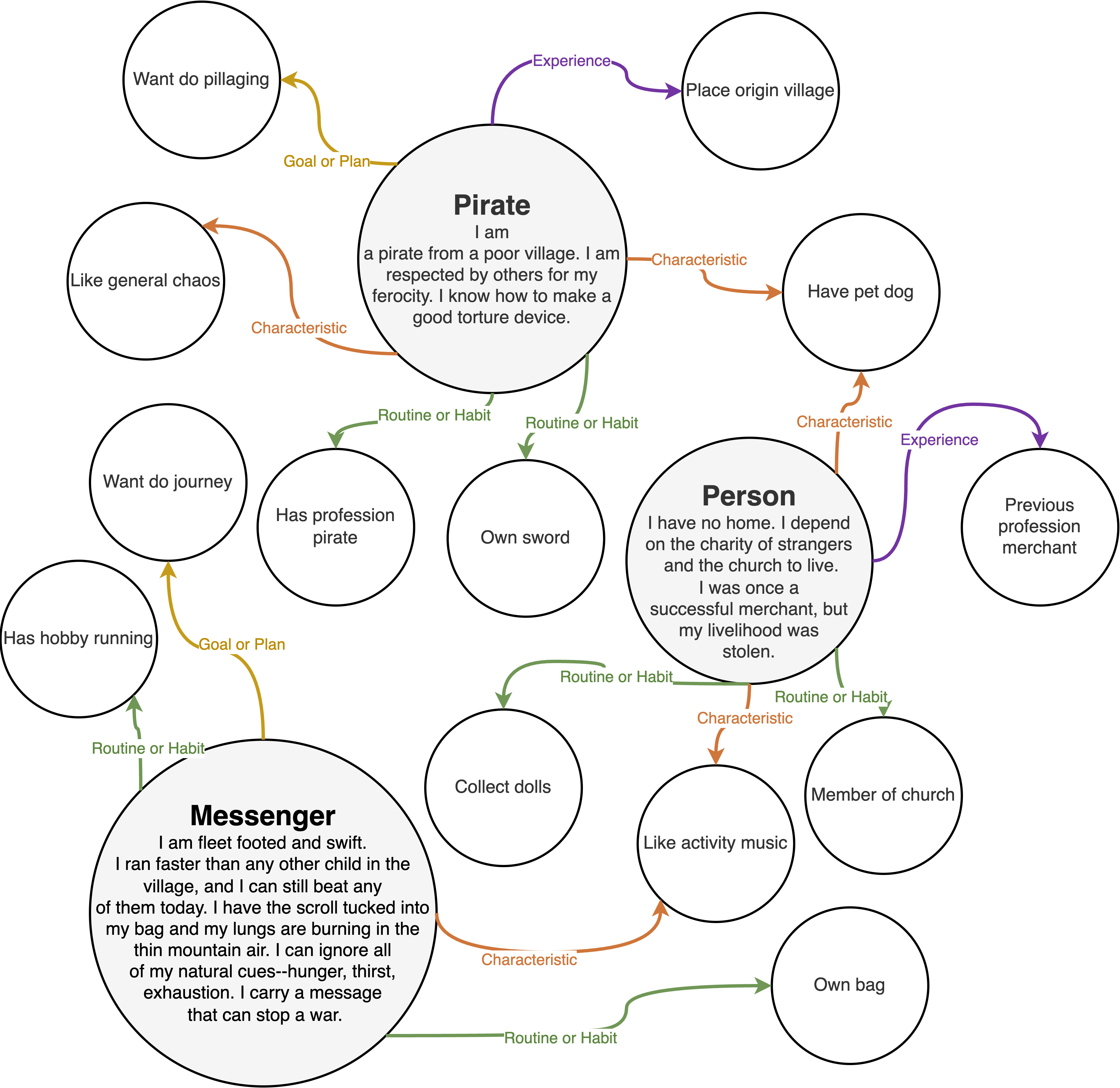}
    \caption{\pck-style persona graph for a few characters from the LIGHT fantasy role-playing dataset. The graph was built from character utterances in conversations.}
    \label{fig:light_ex_graph}
\end{figure}

Constructing persona for these agents can be time-consuming, but \textit{persona extraction} alleviates this by automatically extracting information about a character given past dialogue utterances \citep{wang-etal-2022-extracting,lu-etal-2022-partner,zhou-etal-2023-learning,zhu-etal-2023-paed}.
The task is a precursor to developing personalized response generation models consistent with its assigned persona.
Manually crafting persona can be arduous, and what a character says can be rich with information.
However, current models and techniques for persona extraction are all trained on the same domain, i.e., casual chit-chat, since the standard dataset for training is based on PersonaChat \citep{zhang-etal-2018-personalizing}.
PersonaChat is considered to exist in the ``real world'' and thus persona extraction models trained on it have difficulty extracting persona in new and distant narrative domains such as the fantasy world in LIGHT \citep{urbanek-etal-2019-learning}.

Since data annotation for fine-tuning is costly, we explore post-hoc methods on a trained model to mitigate domain adaptation issues.
Our approach to this domain adaptation issue is to cast it as a natural language inference task, since we want to ensure that extracted persona information is reasonable given the original utterance.
We start with a new model trained on PersonaExt \citep{zhu-etal-2023-paed}, a semi-automatic labeled dataset for persona extraction from PersonaChat utterances.
PersonaExt contains fine-grained relations that do not apply to all narrative settings (``domains''), so we categorize the 105 persona relation types into 4 that are general enough to work with various narratives and characters in varying domains: \textbf{experience}, \textbf{goal or plan}, \textbf{routine or habit}, and \textbf{characteristic}.
For example, both a pirate in a fantasy world and a real-world accountant have persona knowledge about routines and goals, e.g., (I, goal or plan, want to pillage) and (I, goal or plan, want a raise), respectively (\Cref{fig:light_ex_graph}).
These relation types are from the only knowledge graph (KG) specifically designed for persona, PeaCoK \citep{gao-etal-2023-peacok}.

For our methods, we cast persona extraction as a sequence-to-sequence (seq2seq) problem, mapping an utterance to a persona ``triplet'' that can be parsed and added to a KG \citep{ni_generative_2022,wang-etal-2022-extracting,zhu-etal-2023-paed}.
An important problem when generating from out-of-domain utterances is model hallucination or generation of low-quality, generic, or incorrect persona information.
Since KGs should be precise, we introduce a \textit{natural-language inference persona pruning} step, with a natural language inference (NLI) model trained specifically for determining if an extracted triplet can be inferred from the utterance.
We explore three approaches: (1) guided decoding to force the model’s output to be in the correct format, (2) generate many and re-rank, and (3) generate many and classify.
We show that this pruning step reduces false positives (extracting persona when there is no persona information) and improves the quality of extracted character persona when compared to the current state-of-the-art model PAED \citep{zhu-etal-2023-paed}.

\enote{mzhao}{i feel it is actually the other way around? a person's persona determines the actually utterrances output from the person}
The converse relationship between utterances and persona has been investigated before, and even used to make persona extraction datasets \citep{wang-etal-2022-extracting}, but the direction we explore allows us to determine if the extracted persona is \textit{reasonable} given the utterance.

% Contributions
\enote{mzhao}{to be updated after wrapping up}
In this work, we contribute the following:\footnote{All code and models will be released upon publication.}
\begin{itemize}[noitemsep]
    \item \textbf{PersonaExt-PeaCoK}: a semi-automatically adapted dataset PersonaExt \citep{zhu-etal-2023-paed} for training PeaCoK-compatible persona extraction models from dialogue utterances
    \item A trained persona extraction model for extracting PeaCoK relations (experience, goal or plan, routine or habit, characteristic) from dialogue history for the purpose of expanding and adapting PeaCoK to specific narratives
    \item \textbf{Persona-NLI}: an NLI model for evaluating the entailment relationship between a dialogue utterance and extracted persona
    \item Qualitative analysis of model performance on fantasy dataset LIGHT \citep{urbanek-etal-2019-learning}
\end{itemize}

%%%%%%%%%%%%%%%%%%%%%%%%
% Related Work
%%%%%%%%%%%%%%%%%%%%%%%%
\section{Related Work}
\label{sec:related_work}
Our work is relevant in the focus areas of persona-grounded dialogue, persona extraction, and dialogue-specific natural language inference (Dialogue-NLI).
\enote{mzhao}{I think having a separate paragraph of introducing Peacok and Comfact would be helpful.}

\paragraph{Persona-grounded Dialogue}
The primary issues with dialogue agents are their boring and generic responses \citep{li-etal-2016-diversity,khayrallah-sedoc-2021-measuring} and their inability to maintain a consistent persona, often contradicting themselves in conversations \citep{shuster-etal-2022-state}.
An early, widely-used work to address this issue is PersonaChat \citep{zhang-etal-2018-personalizing} which is a manually created dataset of crowdworker conversations paired with 3-4 sentence persona descriptions.
Similarly, Learning in Interactive Games with Humans and Text (LIGHT; \citet{urbanek-etal-2019-learning}) is a dataset created the same way and modeled after PersonaChat, but is set in a fantasy world with fairies, pirates, and kings.
While other fantasy dialogue datasets exist, either they contain prose instead of dialogue \citep{zhu-etal-2023-fireball} and/or do not have structured persona information paired with each character \citep{callison-burch-etal-2022-dungeons,weir_ontologically_2023,van_stegeren_fantastic_2020} (see Appendix \Cref{tab:narrative_datasets} for details).

A missing component of persona-grounded datasets is \textit{commonsense} grounding. 
\citet{gao-etal-2023-peacok} address this issue with PErsonA-grounded COmmonsense Knowledge graph (Pea-CoK) which formalizes relations within and between personas.
Personas are represented as graph nodes consisting of head (subject) and tail (verb phrase) entities connected by set labeled edges (i.e., relations), e.g, (I am a famous pianist, characteristic, skilled in playing the piano, see \Cref{fig:light_ex_graph}).

\paragraph{Persona Extraction}
While the task of persona extraction is new, methods for this task draw from the established field of relation extraction.
In this work, we extract pre-defined relations from a character's utterances to build/expand a persona knowledge graph.

\citet{wang-etal-2022-extracting} separate persona extraction into extraction and inference subtasks, or whether or not the relation can be found verbatim in the utterance. 
They trained GPT-2 Small on Dialog-NLI \citep{welleck-etal-2019-dialogue} and similar to our method, implemented constrained decoding to ensure proper relation format.
While they also trained a separate model for re-ranking generated output, it was not fine-tuned for the NLI task like ours.

\citet{ni_generative_2022,zhu-etal-2023-paed} also cast persona extraction as a seq2seq mapping from dialogue utterance to a persona triplet, e.g., (subject/``head'', relation type, object/``tail'').
\citet{zhu-etal-2023-paed} used a variational auto-encoder (VAE) to create new synthetic examples to train the model to be able to distinguish between related relations (``like" and ``dislike").
They introduced the BART-based zero-shot model PAED, and also released a modified Persona Extraction dataset (PersonaExt) built from the persona extraction dataset from \citet{wang-etal-2022-extracting}.
Similar to \citet{wang-etal-2022-extracting} and us, they use constrained decoding to ensure compliance with expected output format.

Related to persona extraction is persona \textit{expansion}, where commonsense KGs are leveraged to add more information given a presented persona (typically from PersonaChat) \citep{kim-etal-2023-concept,liu_improving_2022} and character identification, where a speaker is identified given their persona information and/or past dialogue \citep{sang-etal-2022-tvshowguess}.

In summary, while we also use seq2seq modeling for persona extraction, we introduce NLI-based reranking and filtering for quality control in a new narrative setting (from ``real world'' to fantasy), and use PeaCoK \citep{gao-etal-2023-peacok} persona relations.

\paragraph{Persona NLI}
Most NLI datasets are not in the narrative or dialogue domain but Dialog-NLI addresses this issue \citet{welleck-etal-2019-dialogue}.
This dataset consists of manually annotated (persona sentence, character utterance) pairs for their entailment relationship from the aforementioned PersonaChat.

We incorporate a newly trained NLI model fine-tuned for persona extraction from dialogue (Persona-NLI).
Since Dialog-NLI is built from PersonaChat, which is also what PersonaExt was built from, we do not include it to train the Persona-NLI model due to test leakage concerns. 
Instead, we turn to a related dataset built for pairing sentences and facts.
Commonsense Fact linking dataset (ComFact) \citep{gao-etal-2022-comfact} is a semi-automatically created dataset for linking statements and commonsense facts from Atomic$_{20}^{20}$ \citep{hwang_comet-_2021}.
More details of adapting ComFact to the NLI task are in \Cref{sec:persona-nli}.

\citet{ammanabrolu-etal-2021-motivate} introduced a commonsense knowledge graph for LIGHT (ATOMIC-LIGHT), but we do not include this data since it does not contain dialogue and our experimental setting assumes no narrative-specific information. 

%%%%%%%%%%%%%%%%%%%%%%%%
% Persona Extraction
%%%%%%%%%%%%%%%%%%%%%%%%
\begin{figure*}
    \centering
\includegraphics[width=\textwidth]{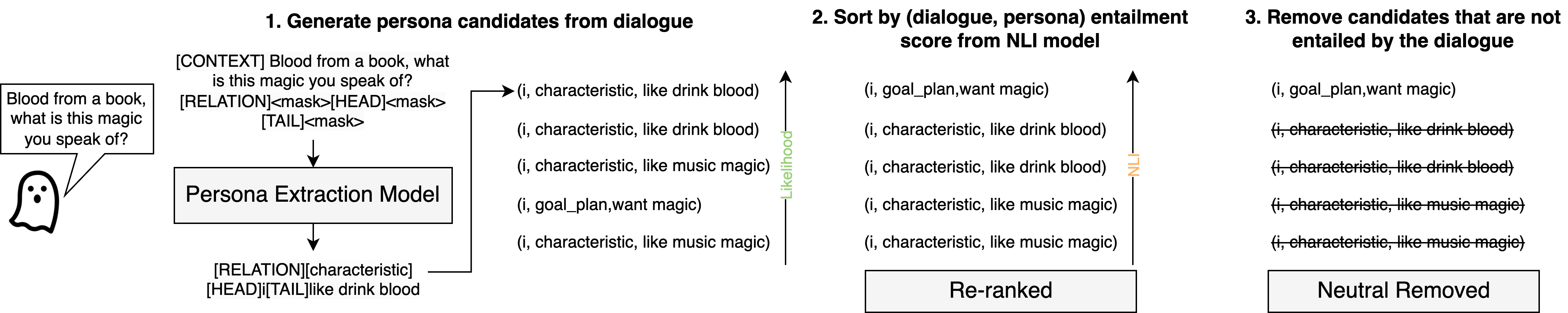}
    \label{fig:method_overview}
    \caption{Overview of persona extraction process from dialogue with a trained seq2seq (BART) and natural language inference (NLI) model. The head, tail, and relation types are parsed from the model output. As seen from the selected example, the proposed NLI re-ranking step correctly adjusts the order of possible extracted persona, penalizing the non-entailed ``(i, characteristic, like drink blood)''.}
\end{figure*}

\section{Persona Extraction from Dialogue}
\label{sec:extract}
Our goal is to automatically adapt a persona KG to a new narrative, starting from character dialogue. 
\pck \citep{gao-etal-2023-peacok} is the first and only KG for personas, and its relation types (i.e, edges) are general enough to fit any narrative setting (see \Cref{fig:light_ex_graph}).
However, \pck was not built from dialogue utterances, so we use PersonaExt \citep{zhu-etal-2023-paed}, a dataset of (utterance, persona relation) pairs to train a \pck-persona extraction model.\footnote{We use PersonaExt instead of the precursor dataset introduced by \citet{wang-etal-2022-extracting} because \citet{zhu-etal-2023-paed} improved upon the annotations and labels.}
We discuss the modification of PersonaExt to fit the \pck persona format in \Cref{sec:data-ppe}.

For the persona extraction task, we fine-tune existing models with a variety of prompts and test different decoding methods. We compare our models to the persona attribute extraction in dialogues (PAED) model
\citep{zhu-etal-2023-paed}.

\subsection{Dataset: PersonaExt-PeaCoK}
\label{sec:data-ppe}
\pck is a KG consisting of head and tail entities with relation types of persona commonsenses, e.g, (``I am a famous pianist'', ``characteristic'', ``skilled in playing the piano''), and does not contain utterances.
While the heads and tails are phrases, combining them into full sentences would lead to repetitive and unnatural utterances.
Instead, we start from an annotated dialogue dataset, \pxt, and semi-automatically convert them into \pck format by re-annotating the dialogue utterances on the \textit{relation level} to \pck relations. We refer to the relabeled \pxt dataset as \textbf{\pxt-\pck}.
The train, validation, and test sets for model training were stratified-split according to the labels.
\Cref{app:data-ppe} reports more details.

\subsection{Dataset: LIGHT}
\label{sec:data-light}
As discussed in \Cref{sec:related_work}, we select LIGHT \citep{urbanek-etal-2019-learning} as the new narrative dataset for evaluation.
We take the best-performing persona extraction model from \Cref{sec:extract} and evaluate on LIGHT.
Unlike \pxt-\pck, there are no ground-truth dialogue-level persona annotations for LIGHT, so we provide a qualitative analysis alongside other intrinsic metrics instead of reporting accuracy metrics.

We download the LIGHT dataset from ParlAI \citep{miller2017parlai}.\footnote{\url{https://github.com/facebookresearch/ParlAI/tree/main/projects/light}}
We found 10,268 dialogues (230 more than \citet{urbanek-etal-2019-learning}) and 1,382 unique characters (e.g., ``a baby dragon'').
``Unique'' characters were determined by counting the unique (character, description) pairs.
We ignore the ``objects'', ``actions'', ``emotes'', and ``actions'' portions of LIGHT since our persona extraction model is based only on dialogue.
We compare the persona extracted from the character description (i.e., the persona profile of a character)\footnote{Persona descriptions are a paragraph where each sentence contains a character trait, we parse it into individual sentences.} to persona extracted from their dialogue utterances.
There are roughly 3-4 persona description sentences per character.
Although a single dialogue utterance, or conversation turn, can consist of more than one sentence, we do not separate into individual sentences.

\subsection{Methods}
\label{subsec:extract-methods}
For the persona extraction model, We fine-tuned the HuggingFace implementation of BART-Large \citep{lewis-etal-2020-bart,wolf-etal-2020-transformers} on \pxt-\pck; \Cref{app:train-details} reports training details.

We created a structured input and output template based on the one used by PAED \citep{zhu-etal-2023-paed}, shown in \Cref{tab:model_template}.
Note that the template tokens are newly added to the model vocabulary with trainable parameters in the embedding lookup layer.
After experimenting with other templates, we found that using special tokens for both the entity markers and relation types led to a better-performing model.
Also, unlike the PAED template where the relation type is generated last, we found better performance when the template is ordered: relation, head entity, and then tail entity.
Other work has used standard relation triplet ordering of head, relation, and tail \citep{wang-etal-2022-extracting}.
The benefit of the relation token in the beginning of the sequence is that the model can then be used to easily generate different relations from the same utterance by changing the relation type.
We initialized embeddings of the added tokens to the average embedding of a short text description (\Cref{tab:added_tokens}).

\begin{table}[]
    \centering
    \resizebox{\columnwidth}{!}{\begin{tabular}{l|l}
        \toprule
        Token & Description \\
        \midrule
       \brak{CONTEXT} & context \\
       \brak{RELATION} & relation \\
       \brak{HEAD} & head entity \\
       \brak{TAIL} & tail entity \\
       \brak{characteristic} & character trait \\
       \brak{no\_relation} & no relation \\
       \brak{routine\_habit} & regularly or consistently do \\
       \brak{goal\_plan} & will do or achieve in the future \\
       \brak{experience} & did in the past \\
       \bottomrule
    \end{tabular}}
    \caption{Tokens added to BART vocabulary for training the persona extraction model. The descriptions for the relation types are from \pck \citep{gao-etal-2023-peacok}. \Cref{tab:model_template} shows an example model input and output.}
    \label{tab:added_tokens}
\end{table}

Since the output needs to be in the correct format for parsing into a triplet, we impose constraints on the generation to ensure template adherence.
These constraints are flexible and can be combined with any decoding method. 

We compare generations from greedy search, beam search, and diverse beam search \citep{vijayakumar_diverse_2018}.
For beam search, we set the number of beams to 5 and return all 5 sequences, with only the most likely sequence presented as the final output.
For diverse beam search, we set the number of beam groups to 5 to thoroughly explore the search space and diversity strength $\lambda=0.4$.

We trained the model on four 48GB NVIDIA RTX A6000 GPUs for 50 epochs, 128 batch size, AdamW optimizer, and $5e-5$ learning rate. 
Training time was 1.5 hours.

\begin{table*}[]
    \centering
    \begin{tabular}{ll}
    \toprule
    Utterance & i am a tour guide at a museum . what do you do for a living ? \\
    \midrule
    Input & \fitcol{\brak{CONTEXT} i am a tour guide at a museum . what do you do for a living ? \brak{RELATION} <MASK> \brak{HEAD} <MASK> \brak{TAIL} <MASK>} \\
    \midrule
    Output & \brak{RELATION} \brak{routine\_habit} \brak{HEAD} i \brak{TAIL} has profession tour guide \\
    \bottomrule
    \end{tabular}
    \caption{The template for model input and output for BART-Large fine-tuned for persona extraction on \pck-\pxt. <MASK> represents BART's mask token, while the tokens in square brackets are special tokens added to the vocabulary. The relation types (e.g., routine\_habit) are also in the model's vocabulary.}
    \label{tab:model_template}
\end{table*}

\subsection{Comparative Models}
\label{subsec:extract-other-models}
We compare our persona extraction model to PAED and ablated versions of our model. 

\paragraph{PAED (zero-shot)}
PAED (see \Cref{sec:related_work}) is another fine-tuned BART model for persona extraction in zero-shot settings,
and we test it with \pck relations.
We trained the PAED model from the author-provided codebase in the 10 unseen label setting.

\paragraph{PAED (fine-tuned)}
While PAED was designed as a zero-shot model to adapt to new relation types, we also include a fully fine-tuned version that is trained on the same train split of \pep as our model.
We trained the PAED model from scratch on PersonaExt-PeaCoK with the same settings as \citet{zhu-etal-2023-paed} (see \Cref{app:paed_reproduce}).

\subsection{Evaluation}
\label{subsec:extract-eval}
We evaluate persona extraction performance through reference and reference-free (intrinsic) metrics.
\pxt-\pck has ground-truth labels to measure accuracy.
In the reference metrics, the head, tail, and relation are parsed from the structured output and compared against the respective gold-standard entity.
However, the LIGHT dataset (in the new domain of fantasy) does not have ground-truth labels, and we rely on intrinsic and manual evaluations.

\paragraph{Reference Accuracy}
For evaluating the extraction we employ the same accuracy metric as in \citet{zhu-etal-2023-paed}.
While \citet{zhu-etal-2023-paed} only awards credit to the model if the entire (\texttt{head}, \texttt{relation}, \texttt{tail}) triplet is predicted correctly (shown in our results as ``Triplet''), we additionally relax this metric and evaluate the model performance separately on identifying the head, relation, and tail entities from the triplet, similar to \citet{wang-etal-2022-extracting}.

Since there are no ground-truth persona triplets for LIGHT, we developed intrinsic (i.e., non-reference-based) metrics alongside qualitative analysis.

\paragraph{Intrinsic Metrics}
The non-accuracy-based metrics focus on the number of returned persona and unique persona. 
\emph{The metrics are measured for each character, overall, and on the dialogue and description datasets.}

We define utterance \textbf{coverage} as the ratio of extracted persona to the number of utterances. 
If the extracted persona has the ``not a relation'' relation type, it is not considered.

We are curious about how many extracted relations are about the speaking character or other characters.
This \textbf{first person} metric measures the ratio of first-person triplets, where the \texttt{head} is ``I'', ``me'', or ``my'', as compared to all extracted triplets.

Regarding diversity and uniqueness, we want to ensure that the model is not assigning the same character traits to everyone.
We look at the rate of unique head and tail entities, referred to as \textbf{unique head} and \textbf{unique tail}.

\paragraph{Persona ``Recall''}
While there are no ground-truth labels for LIGHT on the utterance-level, we can consider the persona descriptions as ``silver'' labels.
We define \textbf{recall} as, using persona triplets extracted from the provided persona descriptions, the ratio of persona relations successfully recovered from the utterances.
For a consistent reference, we use the persona triplets extracted by the model with greedy decoding and no NLI re-ranking or filtering.  

\paragraph{Human Evaluation}
For the qualitative analysis, two authors compared the extracted persona information from the dialogue to that extracted from the character description, and to the character description itself. 
The annotation descriptions are in \Cref{tab:annotation_instructions}.

\begin{table*}
    \centering
    \begin{tabular}{ll|l}
    \toprule 
    Accepted & Sub-category & Guideline \\
    \midrule
    \multirow[c]{2}{*}{Yes} & Directly & Relates to the persona description \\
    & Reasonable & Character trait given the persona description \\
    \midrule
    \multirow[c]{4}{*}{No} & Contradictory & \fitcolm{Candidate persona goes against the persona description. E.g., persona is a dragon but the relation is (I, routine\_habit, am a high priestess)} \\
    & Unreasonable & Unreasonable trait given character description \\
    & Non-specific relation & E.g., (we, characteristic, like activity it) \\
    & Malformed & Parsed triplet is missing an entity, e.g. the tail \\
    \bottomrule
    \end{tabular} 
    \label{tab:annotation_instructions}
    \caption{Guidelines for the human evaluation of extracted persona information. Evaluators were required to select a sub-category within ``Yes'' and ``No'' providing a reason for the selection.}
\end{table*}

Annotation guidelines were created by one author after qualitatively evaluating 5 random characters (sampled with random seed 13).
Each triplet was annotated against the provided character name and 3-4 sentence description (i.e., persona profile in \Cref{sec:data-light}).
Triplets were annotated individually and not as a whole, e.g., two triplets that are contradictory could still be labeled as "reasonable". 
For example, if the description does not contain information about the character's family, "has sibling sister" and "has no siblings" are both reasonable. 
Also, triplets were annotated based on the present. 
If the description specifies "I am a pirate" and the triplet says "I want to be a pirate" it's labeled as a contradiction since the character already is a pirate. 
Further, some triplets were near-duplicates of each other and were consolidated by merging triplets with first-person heads into one subject "self". 
Also, triplets were lowercased before merging. Note this consolidation was only performed for annotations and thus the number of triplets differs from those present in \Cref{tab:results_light_quantitative}.

\begin{table*}[]
    \begin{tabular}{ll}
    \toprule
    \multicolumn{2}{c}{\begin{minipage}{\textwidth}I silently swoop down into the forest to carry off another. I do what I can to please the high priestess, who is my mother. I feast on the bones of the hunter. I like being scratched behind my ear and laying by the fire.\end{minipage}} \\
    \midrule
    Extracted Persona Triplet & Annotation \\
    \midrule
    (self,   characteristic, like food nutritious) & Yes   – Reasonable\\
    (self,   goal\_plan, want do hurts) & Yes   – Reasonable\\
    (self,   characteristic, like food venison) & Yes   – Reasonable\\
    (my   mother, routine\_habit, has profession high priestess) & Yes   – Directly \\
    (self,   routine\_habit, has profession priestess) & No   - Contradictory \\
    (self,   characteristic, like activity play) & Yes   - Reasonable \\
    \bottomrule
    \end{tabular}
    \caption{Example annotations for extracted persona triplets for the ``a dragon'' character in LIGHT.}
    \label{tab:human_eval_ex}
\end{table*}

%%%%%%%%%%%%%%%%%%%%%%%%
% Post-hoc use of Persona NLI
%%%%%%%%%%%%%%%%%%%%%%%%
\section{NLI for Narrative Adaptation}
\label{sec:persona-nli}
When applying the trained persona extraction model to LIGHT, we noticed that a persona relation is always extracted, regardless of the input statement/utterance.
This is expected because persona extraction training data does not contain negative examples (i.e., statement/utterance without a persona).
As a result, it is important to have an automatic method for filtering ``hallucinated'' persona output from the persona extraction model.

We leverage natural language inference (NLI) -- whether a statement \textit{follows} or can \textit{be inferred from} another statement \citep{maccartney-manning-2008-modeling} -- for filtering out the hallucinated persona.
The relationship between dialogue and a corresponding persona statement differs from the typical statement pairs used in conventional NLI training, which often come from formally written text \citep{bowman-etal-2015-large,williams-etal-2018-broad}.
To alleviate potential issues caused by a domain shift, we fine-tune an existing NLI model on (dialogue) utterance-persona statement pairs.
As discussed in \Cref{sec:related_work}, we use the ComFact dataset instead of Dialog-NLI due to train-test leakage with our training dataset, PersonaExt. 
We refer to this fine-tuned model as ``NLI-Persona'' and the non-fine-tuned version as ``NLI-base''.

\subsection{Persona-NLI Model}
\subsubsection{Persona-NLI Dataset}
ComFact contains many annotations at the conversation- or paragraph-level, including whether a fact is relevant at a specific timestep or context window.
To match our utterance-fact setup, we reduce ComFact to only the current statement and facts that were labeled as ``relevant without context'' (RPA) in \citet{gao-etal-2022-comfact}.
Pairs without relation labels were kept as negative examples.
We also filter non-persona related relation types and limit to HasProperty, CapableOf, Desires, xNeed, xAttr, xEffect, xReact, xWant, xIntent, which were identified as persona-related by \citet{gao-etal-2023-peacok}.
We use the original train/dev/test splits in ComFact.
Since \pxt is derived from PersonaChat, we removed any entries in ComFact Persona-Atomic subset that also occurred in our test split to make sure there is no training-testing overlap.
We also include the training and dev splits of \pck-\pxt and ablate the model training over each data subset (\pck-\pxt, ComFact-Persona, and combined \pck-\pxt + ComFact-Persona).

\subsubsection{Training}
Starting from an existing high-performing NLI model, \texttt{nli-deberta-v3-base} \citep{reimers-gurevych-2019-sentence}
\footnote{\url{https://www.sbert.net/docs/pretrained\_cross-encoders.html\#nli}} we fine-tune on our \cmf-derived Persona-NLI dataset for 5 epochs, batch size of 32, learning rate of $2e-05$, and AdamW optimizer on a single 48GB NVIDIA RTX A6000 GPU. \texttt{nli-deberta-v3-base} was trained on SNLI \citep{bowman-etal-2015-large} and MNLI \citep{williams-etal-2018-broad}.
While MNLI does contain training data in the fiction domain ($\approx$20\%), it is prose from contemporary fiction and not dialogue utterances.

We chose the final Persona-NLI model based on F1 score (binary).
Since we only care about ``entailment'' versus ``no entailment'', we do not penalize the model for incorrectly predicting ``neutral'' versus ``contradiction'', and merge those two labels into a binary task.\footnote{The loss is still measured on all three labels, only the accuracy and F1 metrics are measured on the binary classification.}

\Cref{tab:persona-nli-results} shows the results across training (rows) and testing (columns) on the different versions of Persona-NLI dataset.
The model trained on the ComFact-Persona + PeaCoKPersonaExt dataset performed the best and we refer to it as \textbf{Persona-NLI}.
An important note is while the \pck-\pxt appears to have perfect scores, this dataset is very skewed towards the positive class and never predicts ``no entailment'', thus we use the model trained on the more balanced combined dataset of ComFact-Persona + PeaCoKPersonaExt.
The fine-tuning is needed since the NLI model (``w/o fine-tuning'') does not perform as well, which is expected due to domain adaptation issues.

\begin{table*}
    \resizebox{\textwidth}{!}{\begin{tabular}{lrrrr|rrrr|rrrr}
    \toprule
     & \multicolumn{4}{c}{PeaCoK-PersonaExt} & \multicolumn{4}{c}{ComFact-Persona} & \multicolumn{4}{c}{ComFact-Persona+PeaCoKPersonaExt} \\
     & Acc. & F1 & Precision & Recall & Acc. & F1 & Precision & Recall & Acc. & F1 & Precision & Recall \\
    Persona-NLI Dataset &  &  &  &  &  &  &  &  &  &  &  &  \\
    \midrule
    w/o fine-tuning & 0.51 & 0.38 & 0.51 & 0.54 & 0.81 & 0.55 & 0.55 & 0.56 & 0.67 & 0.66 & 0.70 & 0.67 \\\midrule
    PeaCoK-PersonaExt & \textbf{1.00} & \textbf{1.00} & \textbf{1.00} & \textbf{1.00} & 0.11 & 0.10 & 0.50 & 0.50 & 0.52 & 0.38 & 0.74 & 0.52 \\
    \hspace{6pt}+ComFact-Persona & \textbf{1.00} & \textbf{1.00} & \textbf{1.00} & \textbf{1.00} & \textbf{0.89} & 0.75 & \textbf{0.72} & 0.79 & \textbf{0.94} & \textbf{0.94} & \textbf{0.94} & \textbf{0.94} \\
    \bottomrule
    \end{tabular}}
    \caption{Results for NLI-Base model fine-tuned and evaluated on three persona data subsets. The base ``w/o fine-tuning'' model is not trained and only evaluated. F1 metrics are macro-F1. The model fine-tuned on ComFact-Persona+\pck-\pxt performs the best. Accuracy is shortened to ``Acc.'' for space. The 100\% accuracy is due to the large class imbalance in \pep, which is mostly positive examples.}
    \label{tab:persona-nli-results}
\end{table*}

\subsection{NLI-Reranking}
As shown in \Cref{fig:method_overview}, we can adjust the model's final generated output by first generating multiple candidates, scoring each candidate with the Persona-NLI model, and then selecting the one with the new highest score. 
The NLI score for each candidate is based on the entailment score of (utterance, candidate triplet) pairs.

Since Persona-NLI expects a sentence and not a triplet, we provide each candidate as a sentence by concatenating the head and tail, e.g., ``i want magic'' (\Cref{fig:method_overview}).
If the pair is determined to have an entailment relationship, we then adjust the final score of the candidate as the language model score (average token log probability) plus the log probability of entailment.
If there is no entailment relation, the score is kept the same, i.e., only the language model score.
This new score is designed to promote candidates that have high entailment scores, but not penalize the candidates determined to be highly likely by the language model.

This method can only be used with a model that can output multiple candidates, such as BART with a beam search or sampling-based decoding methods.
We refer to this as NLI-\textbf{re-ranking}.

\subsection{NLI Classification}
A stricter use of the NLI model is to completely remove any candidates that cannot be entailed by the utterance, i.e., has a neutral or contradictory relationship.
Since there is no re-ranking, it can be used by models that only generate one candidate.
We refer to this as \textbf{Neutral Removed}.

%%%%%%%%%%%%%%%%%%%%%%%%
% Results
%%%%%%%%%%%%%%%%%%%%%%%%
\section{Persona Extraction Results}
\label{sec:results}
We first evaluate the extraction model performance on the \pck-\pxt dataset and then evaluate the best-performing models on the new narrative setting of LIGHT.

\subsection{PersonaExt-PeaCoK (``In-domain'')}
\label{subsec:extract-results}

\begin{table*}
   \centering
    \resizebox{\textwidth}{!}{\begin{tabular}{llrrrr|r|rr|r}
    \toprule
     &  & \multicolumn{4}{c|}{Label} & Head & \multicolumn{2}{c|}{Tail} & Triplet \\
     &  & Precision & Recall & F1 & Acc. & Acc. & Overlap & Acc. & Acc. \\
    \midrule
    \multirow[c]{3}{*}{Our Model} & Greedy Search  & 0.85 & 0.84 & 0.84 & 0.84 & 0.96 & 0.78 & 0.67 & 0.66 \\
    & Beam Search & 0.85 & 0.83 & 0.84 & 0.83 & 0.96 & 0.78 & 0.67 & 0.66 \\
    & Diverse Beam Search & 0.85 & 0.83 & 0.83 & 0.83 & 0.96 & 0.78 & 0.67 & 0.66 \\
    \midrule
    \multirow[c]{2}{*}{PAED} & (fine-tuned) & 0.83 & 0.83 & 0.83 & 0.83 & 0.95 & 0.75 & 0.62 & 0.61 \\
    & (zero-shot) & 0.52 & 0.06 & 0.03 & 0.06 & 0.08 & 0.07 & 0.07 & 0.06 \\
    \bottomrule
    \end{tabular}}
    \caption{Results on the ``in-domain'' dataset, PersonaExt-PeaCoK. Evaluated on the test split. Accuracy is abbreviated to ``Acc''.}
    \label{tab:results_personaext_peacok}
\end{table*}

The summarized results are shown in \Cref{tab:results_personaext_peacok} (see \Cref{app:results_pck} for detailed results).
We do not include the Persona-NLI re-ranking and removal since those are for adapting the model to the new narrative setting.
The takeaways are as follows:

\paragraph{Our model and PAED perform similarly.}
While PAED is the best-performing model, with an overall triplet accuracy of 0.61, our model is close behind with an accuracy of 0.60. 
PAED in its original zero-shot setting performs poorly. 
Upon further examination, this is due to the ``not'' label having the highest likelihood compared to the other labels.
The original PersonaExt relation types were detailed (e.g., ``school\_status''), so perhaps the model would perform better with more explicitly named relations (e.g., ``attribute\_of'' instead of ``characteristic'').

\paragraph{Relation-label accuracy is impacted by training prevalence.}
As seen in the per-label scores (\Cref{tab:results_persona_extract_detailed}), all models are best at identifying the ``Characteristic'' and ``Routine or Habit'' relations, which are the most prevalent in the training dataset.
The models are the least accurate at identifying ``Experience'', ``Goal or Plan'', and the ``Not a Relation'' categories.

\paragraph{Predicting head entity is a simple task.}
All models, except PAED (zero-shot) have 0.95+ accuracy on identifying the Head entity in the triplet.
This is because there are only a few options for head entities and the vast majority are ``i'' or ``my'' in the training set.
This could impact the generalizability of the model, which we analyze in \Cref{subsec:extract-narrative-results}.

\paragraph{Triplet accuracy is penalized by tail entity.}
All models perform poorly on predicting the tail entity, with an accuracy of only 0.55-0.62.
This low tail accuracy lowers the overall triplet accuracy, which is mostly accurate for the head entities and relations, with accuracy above 0.97 and 0.81, respectively.
The difficulty in tail predicting is most likely due to the length, which is often several tokes as compared to the 1-3 tokens in the head entity, and not all the tail entity tokens are taken directly from the context.
This is clear from the results shown in Appendix \Cref{tab:results_paed_ppe_nophrase}, where the original PAED model is trained and tested on a version of PersonaExt-PeaCoK with the original tail entities (i.e., no tail phrase).
While the tail entity accuracy is significantly better at 0.8, the overall accuracy remains very similar at 0.63.

\paragraph{Decoding method has no impact on results.}
Our model performs similarly with regard to triplet accuracy across the three evaluated decoding methods of greedy, beam, and diverse beam search.
Our theory is the model is well-trained for the \pck-\pxt dataset and thus the most likely candidates are similar across decoding methods.

\subsection{New Narrative Setting (LIGHT)}
\label{subsec:extract-narrative-results}
As discussed in \Cref{subsec:extract-eval}, there are no ground-truth labels for the LIGHT dataset so we turn to intrinsic and human evaluation instead. 
The results are shown in \Cref{tab:results_light_quantitative,tab:results_light_qualitative_annotation_overall}, respectively.
For human evaluation, we annotated extracted triplets from 10 randomly sampled characters across all the models (5304 generations with 1556 unique relations). 
The overall annotations (i.e., ``Yes'' or ``No'') had an inter-annotator agreement (IAA) of $0.90$ and an IAA of $0.85$ for the detailed annotations as measured by Krippendorff's Alpha.\footnote{\url{https://github.com/LightTag/simpledorff}}
The takeaways are as follows:

\paragraph{NLI Removal has the largest impact.}
Removing non-entailed triplets reduces the number of extracted triplets (i.e., not \brak{no\_relation}) by roughly 90\%. 
This extreme reduction also impacts persona recall and utterance coverage.
There is also a difference between Persona-NLI and the general NLI models, as Persona-NLI setting keeps more candidates.
This is most likely due to Persona-NLI model having more exposure to the format of (utterance, persona) pairs.

\paragraph{Intrinsic metrics show no difference between decoding methods.} 
Similar to the results on \pep, there is little performance difference ($\pm 0.2$) between decoding methods within a setting (e.g., Neutral Removed with NLI model) according to the automated (intrinsic) metrics. 
The differences are more apparent in the human evaluation.

\paragraph{Most extracted persona are about the character.}
Across all models, except PAED, there are high ratios of extracted persona that are first-person ($0.9+$, e.g., head entity is ``I'').
The lowest rates of first-person relations are from the PAED models, with only $67\%$ and $43\%$ of relations being first-person from the retrained and zero-shot versions, respectively.

\paragraph{Greedy Search with removing non-entailed triplets from with Persona-NLI is the best model.}
The differences in extracted persona relations are more apparent with human evaluation than from automatic evaluation alone (\Cref{tab:results_light_qualitative_annotation_overall}). 
The number of extracted relations are different for each method, so we focus on the ratio of accepted (i.e., labeled ``Yes'') relations.
With this metric, using Persona-NLI to remove non-entailed persona generated with greedy search is the best performing, with $68\%$ of extracted relations being accepted.
Interestingly, removing the non-entailed relations hurts performance across all other models, as compared to the base setup (i.e., no re-ranking or removal).

\begin{table*}
\centering
\resizebox{\textwidth}{!}{\begin{tabular}{lllrrrrrr|rrrr}
\toprule
 &  &  & \multicolumn{6}{c|}{Per character} & \multicolumn{4}{c}{Overall} \\
 &  &  & Recall & Cov. & Cov. (unique) & 1st person & Unique persona & N. persona & Cov. & Cov. (unique) & 1st person & N. persona \\
\midrule
\multirow[c]{6}{*}{Neutral Removed} & \multirow[c]{3}{*}{NLI} & Beam Search & 0.02 & 0.09 & 0.08 & 0.87 & 0.90 & 7.55 & 0.09 & 0.05 & 0.92 & 6053 \\
 &  & Diverse Beam Search & 0.02 & 0.08 & 0.07 & 0.84 & 0.87 & 6.46 & 0.08 & 0.04 & 0.91 & 5020 \\
 &  & Greedy Search & 0.02 & 0.06 & 0.05 & 0.83 & 0.82 & 4.71 & 0.06 & 0.03 & 0.94 & 3756 \\\cline{2-13}
 & \multirow[c]{3}{*}{Persona-NLI} & Beam Search & 0.07 & 0.21 & 0.20 & \textbf{0.97} & 0.93 & 17.69 & 0.21 & 0.06 & 0.99 & 7806 \\
 &  & Diverse Beam Search & 0.07 & 0.15 & 0.14 & 0.95 & 0.92 & 12.38 & 0.15 & 0.04 & 0.98 & 5783 \\
 &  & Greedy Search & 0.06 & 0.09 & 0.08 & 0.94 & 0.90 & 7.54 & 0.09 & 0.03 & 4030 & \textbf{0.99} \\
\midrule\multirow[c]{4}{*}{Re-ranking} & \multirow[c]{2}{*}{NLI} & Beam Search & 0.30 & 0.85 & 0.78 & 0.89 & 0.91 & 69.65 & 0.85 & 0.31 & 0.89 & 40331 \\
 &  & Diverse Beam Search & 0.31 & 0.89 & 0.81 & 0.89 & 0.91 & 72.77 & 0.89 & 0.32 & 0.88 & 42469\\
 \cline{2-13}
 & \multirow[c]{2}{*}{Persona-NLI} & Beam Search & 0.29 & 0.87 & 0.79 & 0.90 & 0.92 & 70.93 & 0.87 & 0.31 & 0.89 & 40226 \\
 &  & Diverse Beam Search & 0.31 & 0.89 & \textbf{0.81} & 0.89 & 0.92 & 72.96 & 0.89 & 0.32 & 0.89 & 41866 \\
 \midrule
 \multirow[c]{3}{*}{Base} & & Beam Search & 0.30 & 0.85 & 0.78 & 0.89 & 0.91 & 69.44 & 0.85 & 0.31 & 0.89 & 40100 \\
 &  & Diverse Beam Search & \textbf{0.31} & \textbf{0.89} & 0.81 & 0.89 & 0.91 & 72.57 & \textbf{0.89} & 0.32 & 0.89 & 42288 \\
 &  & Greedy Search & 0.31 & 0.87 & 0.79 & 0.89 & 0.91 & 70.64 & 0.87 & 0.32 & 0.88 & 41709  \\
\midrule\multirow[c]{2}{*}{PAED} & & (fine-tuned) & 0.06 & 0.83 & 0.79 & 0.67 & \textbf{0.96} & 72.99 & 0.83 & \textbf{0.46} & 0.66 & 59787  \\
& & (zeroshot) & 0.00 & 0.03 & 0.03 & 0.43 & 0.71 & 2.77 & 0.03 & 0.02 & 0.57 & 2417 \\
\bottomrule
\end{tabular}}
\caption{Quantitative results with intrinsic metrics across all models on the new narrative setting (fantasy, LIGHT). ``Cov.'' is short for utterance coverage. The Overall metrics are across all extracted persona and not analyzed for each character, thus the persona recall metric is not applicable.}
\label{tab:results_light_quantitative}
\end{table*}

\begin{table*}
\centering
\begin{tabular}{lllrrr}
\toprule
 & NLI Model & Decoding Method & No & Yes & Ratio accepted \\
\midrule
\multirow[c]{6}{*}{Neutral Removed} & \multirow[c]{3}{*}{NLI} & Beam Search & 30 & 38 & 0.56 \\
 &  & Diverse Beam Search & 21 & 32 & 0.60 \\
 &  & Greedy Search & 18 & 27 & 0.60 \\\cline{2-6}
 & \multirow[c]{3}{*}{Persona-NLI} & Beam Search & 75 & 82 & 0.52 \\
 &  & Diverse Beam Search & 49 & 68 & 0.58 \\
 &  & Greedy Search & 23 & 49 & \textbf{0.68} \\
\midrule\multirow[c]{4}{*}{Re-ranking} & \multirow[c]{2}{*}{NLI} & Beam Search & 195 & 332 & 0.63 \\
 &  & Diverse Beam Search & 202 & 334 & 0.62 \\\cline{2-6}
 & \multirow[c]{2}{*}{Persona-NLI} & Beam Search & 209 & 324 & 0.61 \\
 &  & Diverse Beam Search & 206 & 333 & 0.62 \\
 \midrule
 \multirow[c]{3}{*}{Base} & & Beam Search & 193 & 331 & 0.63 \\
 &  & Diverse Beam Search & 202 & 335 & 0.62 \\
 &  & Greedy Search & 194 & 338 & 0.64 \\
\bottomrule
\end{tabular}
\caption{Qualitative, human evaluations for our model evaluated on the new fantasy narrative setting (LIGHT). Inter-annotator agreement (IAA, Krippendorff's Alpha measured with two annotators) of $0.90$. ``Ratio Accepted'' refers to extracted persona annotated with the ``Yes'' category (i.e., marked as appropriate by both annotators).}
\label{tab:results_light_qualitative_annotation_overall}
\end{table*}

%%%%%%%%%%%%%%%%%%%%%%%%
% Discussion
%%%%%%%%%%%%%%%%%%%%%%%%
\section{Discussion}
\label{sec:discussion}
The use-case of our persona extraction model is to adapt a character knowledge graph (\pck) given past dialogue from a narrative setting that differs from the original training data.
We showcase the ability of our best-performing model by building a graph with a few of the manually annotated LIGHT characters from \citet{urbanek-etal-2019-learning}, shown in \Cref{fig:light_ex_graph}. 
Only three of the ten annotated characters with the extracted persona manually determined to be related are shown (\Cref{subsec:extract-narrative-results}).
This graph can then be used for persona-grounded dialogue \citep{gao-etal-2023-peacok}.

From the dialogue, the persona extraction model was able to extract persona information beyond the persona.
For example, while the Pirate obviously (from the description) works as a pirate and is from a village, they also have a pet dog and own a sword. 

Not surprisingly, the model with greedy search performed the best, with regards to manually accepted persona, over the diverse decoding methods of beam and diverse beam search.
This nods to the ``quality vs diversity'' trade-off seen in other areas of text generation.

Another trade-off was ``quality over quantity'', since the Neutral Removed models returned significantly less persona than the Base and Re-ranking models.
This was a benefit since there were less personas to annotate (e.g., an average of 4-20 extracted persona per character instead of 70 (\Cref{tab:results_light_quantitative})).
Since even the best model had an acceptance rate of only 68\%, there is still a need for a human evaluation step.
This is further enforced by the lack of clear relationship between the intrinsic, quantitative metrics and the annotations.
The silver persona ``recall'' metric proved uninformative and ended up having higher correlation with the number of returned persona than an indication of quality.

%%%%%%%%%%%%%%%%%%%%%%%%
% Conclusion
%%%%%%%%%%%%%%%%%%%%%%%%
\section{Conclusion}
\label{sec:conclusion}
Our goal was to address the challenges in adapting a persona extraction model trained on one narrative setting (e.g., real-world ``chit-chat'') to another setting (fantasy).
We modeled persona extraction as a seq2seq problem and fine-tuned BART on PersonaExt, a dataset of (utterance, persona) pairs built from DialogNLI and PersonaChat.
In order to extract persona information applicable to any narrative setting, we converted the ``chit-chat'' PersonaExt to the general relations from PeaCoK (e.g., ``characteristic'').
With our trained BART-based persona extraction model, we evaluated two different post-hoc techniques to extract persona from dialogue from a different narrative setting, the fantasy world of LIGHT.
We experimented with natural language inference (NLI)-based re-ranking and removal of persona candidates, and determined that leveraging inference information to remove persona candidates that can not be inferred from the utterance worked the best according to human evaluation.

%%%%%%%%%%%%%%%%%%%%%%%%
% Ethics & Limitations
%%%%%%%%%%%%%%%%%%%%%%%%
%\clearpage
\section*{Limitations}
\label{sec:limits}
The main limitation of this work is assuming that existing dialogue from a character is available to extract persona information from.
Further, our methods are only evaluated on dialogue utterances and not on other character-related text such as prose (see \Cref{app:related-data}).

\section*{Ethical Considerations}
\label{sec:ethics}
The ethical concerns of this work center on the possibility of automatically impersonating an existing person, rather than the intended use case of fictional characters.
Further, our model is trained on the PersonaExt dataset (derived from the crowdsourced PersonaChat), so we cannot guarantee no presence of offensive language.
Manual evaluation is always the best final step, and we encourage developers who use our method for persona extraction to add a toxicity (e.g., hate or offensive speech) evaluation step in addition to quality evaluation.

%%%%%%%%%%%%%%%%%%%%%%%%
% Acknowledgements
%%%%%%%%%%%%%%%%%%%%%%%%
% \section*{Acknowledgements}

\clearpage
\bibliography{anthology,zotero,custom}
\bibliographystyle{acl_natbib}

%%%%%%%%%%%%%%%%%%%%%%%%
% Appendix
%%%%%%%%%%%%%%%%%%%%%%%%
\clearpage
\appendix
\section{Narrative-Specific Datasets}
\label{app:related-data}
While there are many existing datasets to support narrative analyses, we require a dataset that supports our tasks of \textit{persona extraction from dialogue} and \textit{dialogue agent roleplay}.
The datasets we considered are in \Cref{tab:narrative_datasets}.
The datasets varied in size, type of dialogue, and type of persona description.
The dialogue from a character can either be presented in prose or conversational dialogue form, e.g., ``as the knight slew the dragon he yelled ''for glory!'" versus ``Knight: For glory!".
Also, to evaluate our extracted person graph we need ground-truth persona data in the form of character-level information.
This persona information can either be absent, structured, or unstructured.
Structured persona information is either provided as sentences, as in LIGHT \citep{urbanek-etal-2019-learning}, or in tabular format \citep{li_aloha_2020,zhu-etal-2023-fireball}.

% \usepackage{booktabs}
% \usepackage[normalem]{ulem}
\begin{table*}[]
\centering
\resizebox{\textwidth}{!}{
\begin{tabular}{@{}lllllll@{}}
\toprule
Name & Source of Data & World Setting & N. characters & Dialogue? & Persona? \\
\midrule
Dungeons and Dragons \citep{callison-burch-etal-2022-dungeons} & Play-By-Post & Dungeons and Dragons & 7168 (est.) & prose & No \\
LIGHT \citep{urbanek-etal-2019-learning} & LIGHT crowdsource platform & LIGHT & 1755 & dialogue & structured \\
TorchLight II \citep{van_stegeren_fantastic_2020} & Torchlight II & Torchlight & 82 & dialogue & No \\
Fireball \citep{zhu-etal-2023-fireball} & Avrae bot on Discord & Dungeons and Dragons & 160K & prose & unstructured \\
KNUDGE \citep{weir_ontologically_2023} & The Outer Worlds video game & The Outer Worlds video game & 81 & dialogue & unstructured \\
Storium \citep{akoury-etal-2020-storium} & Storium & Multiple & 25,955 & prose & unstructured \\
TVShowGuess \citep{sang-etal-2022-tvshowguess} & TVMegaSite & Multiple &  & dialogue & No \\
Star Wars: KOTOR \citep{van_stegeren_fantastic_2020} & Star Wars: KOTOR & Star Wars universe & 556 & dialogue & No \\
HLA-Chat \citep{li_aloha_2020} & TV Tropes & Multiple & 45,821 & dialogue & structured \\
LiSCU \citep{brahman-etal-2021-characters-tell} & schmoop, SparkNotes, cliffNotes, LitCharts & Multiple & 9,499 & prose & no
\\\bottomrule
\end{tabular}}
\caption{Datasets from the literature considered in this work for persona knowledge graph narrative adaptation. LIGHT stands for Learning in Interactive Games with Humans and Text. KOTOR stands for  Knights of the Old Republic. LiSCU stands for Literature Summary and   Character Understanding.}
\label{tab:narrative_datasets}
\end{table*}

\section{\pep Dataset Details}
\label{app:data-ppe}
We modified the \pxt dataset \citep{zhu-etal-2023-paed} for compatibility with \pck \citep{gao-etal-2023-peacok} by re-labeling the relations and converting the single-word tails to phrases.
We refer to the relabeled \pxt dataset as \textbf{\pep}.
The train, validation, and test sets for model training were stratified-split according to the new labels (see \Cref{tab:peacok_personaex_data} for details.)

\paragraph{Conversion to \pck Labels}
The 105 \pxt labels were mapped to one of four \pck labels: routine or habit, characteristic, goal or plan, and experience.
While there is also a ``relationship'' label, this is a meta-label that can occur between attributes \textit{across} characters, so we do not include it as a label for the extractor.
Two authors manually mapped the labels and disagreements were discussed until annotations were unanimous.
The relation mappings are shown in \Cref{tab:personaext_mapping}.

\paragraph{Tail Phrase Creation}
To create the tail phrases we combine a \pxt relation with the tail, e.g., ``(have\_family, wife)'' becomes ``have family wife.''
This can create phrases which are not proper English and we leave addressing this to future experiments.

\begin{table*}[]
    \centering
    \resizebox{\textwidth}{!}{\begin{tabular}{lcc}
    \toprule
    \multicolumn{1}{c}{Utterance} & PersonaExt Triplet & PeaCoK Triplet \\
    \midrule
    \resizecol{0.4}{clothes . i   am going to auburn for med school .} & (i, attend\_school, auburn) & (i, routine\_habit, attend school auburn) \\
    \midrule
    \resizecol{0.4}{lol , maybe . i usually wear band shirts and   ruffle sleeves , skinny jeans and leggings} & (i, favorite, ruffle sleeves) & (i,   characteristic, favorite ruffle sleeves) \\
    \midrule
    \resizecol{0.4}{hmmm . i would travel more if i had the money . you travel or   sing ?} & (i, want, money) & (i, goal\_plan,   want money) \\
    \midrule
    \resizecol{0.4}{i never flew a plane , but i have flow from france to canada where i live} & (i, place\_origin, france) & (i,   experience, place origin france) \\
    \midrule
    \resizecol{0.4}{do you have friends here , i have lots here . You going back   to school ?} & (i, misc\_attribute, friends) & (i,   not, misc   attribute friends) \\
    \bottomrule
    \end{tabular}}
        \caption{Example datapoints from PersonaExt modified to fit PeaCoK format.}
    \label{tab:personaext_peacok_map}
\end{table*}

\begin{table*}
\resizebox{\textwidth}{!}{\begin{tabular}{ll}
\toprule
Relation & PersonaExt Relations \\
\midrule
Characteristic & \fullcol{belief, favorite, favorite\_activity, favorite\_animal, favorite\_book, favorite\_color, favorite\_drink, favorite\_food, favorite\_hobby, favorite\_movie, favorite\_music, favorite\_music\_artist, favorite\_place, favorite\_season, favorite\_show, favorite\_sport, gender, has\_ability, has\_age, have\_chidren, have\_children, have\_family, have\_no, have\_no\_children, have\_no\_family, have\_no\_sibling, have\_pet, have\_sibling, have\_vehicle, health\_status, like\_activity, like\_animal, like\_character, like\_color, like\_drink, like\_food, like\_general, like\_goto, like\_movie, like\_music, like\_read, like\_sport, like\_sports, like\_subject, like\_watching, name, physical\_attribute, race, scared\_of, sexual\_orientation, weakness} \\
\midrule
Experience & \fullcol{has\_degree, international\_exp, nationality, place\_origin, pre\_employed\_by\_company, pre\_employed\_by\_general, previous\_profession, raised\_by, used\_to} \\
\midrule
Goal\ plan & \fullcol{want, want\_do, want\_job, want\_no} \\
\midrule
Not & \fullcol{fall\_out, have, industry, misc\_attribute, other, work\_schedule} \\
\midrule
Routine\ habit & \fullcol{allergy\_to, attend\_school, collect, diet, dislike, dislike\_activity, dislike\_animal, dislike\_color, dislike\_drink, dislike\_food, dislike\_job, dislike\_music, dislike\_sport, dislike\_subject, do\_not\_do, do\_not\_drink, do\_not\_eat, doing, employed\_by\_company, employed\_by\_general, get\_along, has\_hobby, has\_profession, job\_status, live\_in\_citystatecountry, live\_in\_general, live\_with, marital\_status, member\_of, never\_do, own, relationship, school\_status, teach, worry\_about} \\
\bottomrule
\end{tabular}}
\caption{Corresponding PeaCoK relation type for each of the 105 PersonaExt relations. The 'not' category refers to not being mapped to a PeaCoK relation type.}
\label{tab:personaext_mapping}
\end{table*}

\begin{table*}
    \centering
    \begin{tabular}{lrrr|r}
        \toprule
        & Train & Dev & Test & Total \\
        \midrule
        Characteristc & 0.57 & 0.57 & 0.57 & 0.57 \\
        Experience & 0.03 & 0.03 & 0.03 & 0.03 \\
        Goal or Plan & 0.04 & 0.04 & 0.04 & 0.04 \\
        Routine or Habit & 0.31 & 0.31 & 0.31 & 0.31 \\
        Not a Relation & 0.04 & 0.04 & 0.04 & 0.04 \\
        \midrule
        All & 28412 & 3157 & 3508 & 35077 \\
        \bottomrule
    \end{tabular}
    \caption{Size and label distribution for each split in the PeaCoK-PersonaExt dataset. The Characteristic and Routine or Habit attributes are the most frequent. The train, validation, and test sets for model training were stratified-split according to the labels.}
    \label{tab:peacok_personaex_data}
\end{table*}

\begin{table*}
    \centering
    \resizebox{\textwidth}{!}{\begin{tabular}{ll}
        \toprule
        & Top Entity Values\\
        \midrule
        Head & \fullcol{i (2920), my (309), me (90), we (33), my dad (27), my mom (25), mine (17), my mother (17), my parents (16), my brother (8), my father (7), dog (5), parents (4), my wife (4), my boyfriend (4), my family (4), family (3), my husband (2), best friend (2), friends (1), our (1), friend (1), mom (1), my girlfriend (1), us (1), cat (1), my dads (1), my son (1), my grandma (1), my best friend (1)} \\
        \midrule
        Tail & \fullcol{have pet dog (39), have pet cat (37), favorite food pizza (24), diet vegan (21), have pet cats (18), like animal dogs (18), marital status married (18), like music country (17), has profession teacher (17), have friends (17), attend school high school (17), like activity hiking (15), has profession nurse (15), like activity video games (15), have vehicle car (14), like activity shopping (14), have pet dogs (14), like animal dog (13), like color blue (13), like activity travel (13), has profession artist (11), have sibling sister (11), physical attribute short (11), like sports basketball (10), like goto beach (10), physical attribute tall (10), like drink coffee (10), like activity reading (10), place origin farm (10)} \\
        \bottomrule
    \end{tabular}}
    \caption{The most prevalent values for the Head and Tail entities in the PeaCoK-PersonaExt dataset. Only the tail entities with at least 10 occurances are shown. There are 30 unique head entitites and 1708 unique tail entities.}
    \label{tab:ppe-entity}
\end{table*}

%%%%%%%%%%%%%%%%%%
% Model Training
%%%%%%%%%%%%%%%%%%%%%%%%%%%%
\section{Persona Extraction Model Training Details}
\label{app:train-details}
We fine-tuned BART \citep{lewis-etal-2020-bart} with the HuggingFace Seq2SeqTrainer \citep{wolf-etal-2020-transformers}. 
The hyperparameter settings are shown in \Cref{tab:model-params}.

\subsection{BART Fine-tuning Templates}
We experimented with variations of the structured input and output templates for fine-tuning BART.
The templates are shown in \Cref{tab:train_templates} and the results are in \Cref{tab:results_persona_extract_detailed} (summarized in \Cref{tab:results_persona_extract_templates}).

Our models perform similarly on the overall accuracy (i.e., triplet accuracy) regardless of the input or output template, other than Relation-first, with scores around $0.55$. 
The model with the Relation-first template had a higher overall accuracy score of $0.60$, which is comparable to the score from comparative model PAED ($0.61$).
We use the Relation-first template for all other experiments.

\begin{table*}
   \centering
    \begin{tabular}{llr}
        \toprule
        Param. & & Value \\
        \toprule
        Batch size & & 32 \\
        Epochs & & 20 \\
        Seed & & 42 \\ \midrule
        Optimizer & & AdamW \\
        	& Learning Rate & $5e-5$ \\
        & Scheduler & linear \\
        & Warmup & 1 epoch \\
        & weight decay & 0 \\
        & adam\_beta1 & 0.9 \\
  	& adam\_beta2 & 0.999 \\
  	& adam\_epsilon & $1e-8$ \\
        \bottomrule
    \end{tabular}
    \caption{Hyperparameter settings for the persona extraction model. The best model was saved at the end according to evaluation loss on the validation set. Model was trained on 4 GPUs, batch size shown is the effective batch size.}
    \label{tab:model-params}
\end{table*}

\begin{table*}
\resizebox{\textwidth}{!}{\begin{tabular}{lll}
\toprule
\multicolumn{1}{c}{Name} & \multicolumn{1}{c}{Input} & \multicolumn{1}{c}{Output}  \\
\midrule
PAED & \resizecol{0.4}{Context : \{context\} Head Entity : <mask>, Tail Entity : <mask> , Relation : <mask> .} & \resizecol{0.4}{Head Entity : \{head\} , Tail Entity : \{tail\} , Relation : \{relation\} .} \\
\midrule
Tokens & \resizecol{0.4}{\brak{CONTEXT} \{context\} \brak{HEAD} <mask> \brak{TAIL} <mask> \brak{RELATION} <mask>} & \resizecol{0.4}{\brak{HEAD} \{head\} \brak{TAIL} \{tail\} \brak{RELATION} \{relation\}} \\
\midrule
Relation-first & \resizecol{0.4}{\brak{CONTEXT} {context} \brak{RELATION} <mask> \brak{HEAD} <mask> \brak{TAIL} <mask>} & \resizecol{0.4}{\brak{RELATION} \{relation\} \brak{HEAD} \{head\} \brak{TAIL} \{tail\}} \\
\midrule
Relation-first-nomask & \resizecol{0.4}{\brak{CONTEXT} \{context\}} & \resizecol{0.4}{\brak{RELATION} \{relation\} \brak{HEAD} \{head\} \brak{TAIL} \{tail\}} \\
\bottomrule
\end{tabular}}
\caption{Evaluated input and output prompts for fine-tuning BART for persona extraction task.}
\label{tab:train_templates}
\end{table*}

%%%%%%%%%%%%%%%
% Metrics
%%%%%%%%%%%%%%%
\begin{table*}
    \centering
    \begin{tabular}{@{}llrrrr|r|r|r@{}}
    \toprule
     & & \multicolumn{4}{c}{Relation} & \multicolumn{1}{c}{Head} & \multicolumn{1}{c}{Tail} & \multicolumn{1}{c}{Triplet} \\
     & & \multicolumn{1}{c}{Prec.} & \multicolumn{1}{c}{Recall} & \multicolumn{1}{c}{F1} & \multicolumn{1}{c|}{Acc.} & \multicolumn{1}{c|}{Acc.} & \multicolumn{1}{c|}{Acc.} & \multicolumn{1}{c}{Acc.} \\ \midrule
     \multirow{4}{*}{BART} & Token prompt & 0.81 & 0.81 & 0.81 & 0.81 & 0.97 & 0.56 & 0.55 \\ \cline{2-9}
    & Relation-first & 0.83 & 0.83 & 0.83 & 0.83 & 0.98 & 0.61 & 0.60 \\ \cline{2-9}
    & Relation-first-nomask & 0.81 & 0.81 & 0.81 & 0.81 & 0.98 & 0.55 & 0.55 \\ \cline{2-9}
    & PAED prompt & 0.81 & 0.81 & 0.81 & 0.81 & 0.97 & 0.56 & 0.56 \\
    \midrule
    PAED & & 0.83 & 0.83 & 0.83 & 0.83 & 0.98 & 0.62 & 0.61 \\
    \bottomrule
    \end{tabular}
    \caption{Summary results for our model (BART) trained with the variety of input and output templates (\Cref{tab:train_templates}) and comparative model PAED trained and tested on the PersonaExt-PeaCoK dataset. Results are on the test set.}
    \label{tab:results_persona_extract_templates}
\end{table*}

\begin{table*}
\centering
    \begin{tabular}{lllrrrr|r|r|r}
    \toprule
     & & &  \multicolumn{4}{c}{Relation} & \multicolumn{1}{c}{Head} & \multicolumn{1}{c}{Tail} & \multicolumn{1}{c}{Triplet} \\
     & & & \multicolumn{1}{c}{Prec.} & \multicolumn{1}{c}{Recall} & \multicolumn{1}{c}{F1} & \multicolumn{1}{c|}{Acc.} & \multicolumn{1}{c|}{Acc.} & \multicolumn{1}{c|}{Acc.} & \multicolumn{1}{c}{Acc.} \\ \midrule
\multirow{24}{*}{BART} & \multirow{6}{*}{PAED prompt} & Char. & 0.85 & 0.91 & 0.88 & \multicolumn{1}{r|}{0.91} & \multicolumn{1}{r|}{0.99} & 0.57  & 0.56 \\
 && Exp. & 0.70 & 0.64 & 0.67 & \multicolumn{1}{r|}{0.64} & \multicolumn{1}{r|}{0.91} & 0.54  & 0.50 \\
 && Goal & 0.64 & 0.76 & 0.69 & \multicolumn{1}{r|}{0.76} & \multicolumn{1}{r|}{0.99} & 0.64  & 0.64 \\
 && Routine & 0.79 & 0.70 & 0.75 & \multicolumn{1}{r|}{0.70} & \multicolumn{1}{r|}{0.96} & 0.57  & 0.56 \\
 && Not & 0.69 & 0.54 & 0.61 & \multicolumn{1}{r|}{0.54} & \multicolumn{1}{r|}{0.95} & 0.46  & 0.46 \\ \cline{3-10}
 && Overall & 0.81 & 0.81 & 0.81 & \multicolumn{1}{r|}{0.81} & \multicolumn{1}{r|}{0.97} & 0.56  & 0.56 \\  \cline{2-10}
 
 & \multirow{6}{*}{Tokens} & Char. & 0.84 & 0.91 & 0.88 & \multicolumn{1}{r|}{0.91} & \multicolumn{1}{r|}{0.98} & 0.57 & 0.57 \\
 && Exp. & 0.65 & 0.67 & 0.66 & \multicolumn{1}{r|}{0.67} & \multicolumn{1}{r|}{0.92} & 0.57  & 0.52 \\
 && Goal & 0.66 & 0.73 & 0.70 & \multicolumn{1}{r|}{0.73} & \multicolumn{1}{r|}{0.98} & 0.58  & 0.58 \\
 && Routine & 0.80 & 0.69 & 0.74 & \multicolumn{1}{r|}{0.69} & \multicolumn{1}{r|}{0.97} & 0.54  & 0.54 \\
 && Not & 0.67 & 0.58 & 0.62 & \multicolumn{1}{r|}{0.58} & \multicolumn{1}{r|}{0.96} & 0.49  & 0.49 \\ \cline{3-10}
 && Overall & 0.81 & 0.81 & 0.81 & \multicolumn{1}{r|}{0.81} & \multicolumn{1}{r|}{0.97} & 0.56  & 0.55 \\  \cline{2-10}
 
&  \multirow{6}{*}{Relation-first} & Char. & 0.86 & 0.93 & 0.89 & \multicolumn{1}{r|}{0.93} & \multicolumn{1}{r|}{0.99} & 0.62  & 0.62 \\
 && Exp. & 0.80 & 0.66 & 0.73 & \multicolumn{1}{r|}{0.66} & \multicolumn{1}{r|}{0.93} & 0.61  & 0.58 \\
 && Goal & 0.69 & 0.67 & 0.68 & \multicolumn{1}{r|}{0.67} & \multicolumn{1}{r|}{0.98} & 0.57  & 0.57 \\
 && Routine & 0.80 & 0.74 & 0.77 & \multicolumn{1}{r|}{0.74} & \multicolumn{1}{r|}{0.97} & 0.60  & 0.59 \\
 && Not & 0.84 & 0.55 & 0.66 & \multicolumn{1}{r|}{0.55} & \multicolumn{1}{r|}{0.94} & 0.47  & 0.45 \\ \cline{3-10}
 && Overall & 0.83 & 0.83 & 0.83 & \multicolumn{1}{r|}{0.83} & \multicolumn{1}{r|}{0.98} & 0.61  & 0.60 \\  \cline{2-10}
 
 & \multirow{6}{*}{Relation-first-nomask} & Char. & 0.87 & 0.87 & 0.87 & \multicolumn{1}{r|}{0.87} & \multicolumn{1}{r|}{0.98} & 0.54  & 0.53 \\
 && Exp. & 0.71 & 0.68 & 0.70 & \multicolumn{1}{r|}{0.68} & \multicolumn{1}{r|}{0.93} & 0.58  & 0.53 \\
 && Goal & 0.64 & 0.70 & 0.67 & \multicolumn{1}{r|}{0.70} & \multicolumn{1}{r|}{0.98} & 0.60  & 0.59 \\
 && Routine & 0.75 & 0.76 & 0.75 & \multicolumn{1}{r|}{0.76} & \multicolumn{1}{r|}{0.97} & 0.58  & 0.58 \\
 && Not & 0.70 & 0.58 & 0.64 & \multicolumn{1}{r|}{0.58} & \multicolumn{1}{r|}{0.94} & 0.48  & 0.47 \\ \cline{3-10}
 && Overall & 0.81 & 0.81 & 0.81 & \multicolumn{1}{r|}{0.81} & \multicolumn{1}{r|}{0.98} & 0.55  & 0.55 \\ \midrule
 
\multirow{6}{*}{PAED} && Char. & 0.87 & 0.91 & 0.89 & \multicolumn{1}{r|}{0.91} & \multicolumn{1}{r|}{0.99} & 0.63 & 0.62 \\
 && Exp. & 0.70 & 0.68 & 0.69 & \multicolumn{1}{r|}{0.68} & \multicolumn{1}{r|}{0.92} & 0.65  & 0.60 \\
 && Goal & 0.73 & 0.75 & 0.74 & \multicolumn{1}{r|}{0.75} & \multicolumn{1}{r|}{0.98} & 0.67  & 0.67 \\
 && Routine & 0.81 & 0.74 & 0.78 & \multicolumn{1}{r|}{0.74} & \multicolumn{1}{r|}{0.97} & 0.62  & 0.60 \\
 && Not & 0.69 & 0.65 & 0.67 & \multicolumn{1}{r|}{0.65} & \multicolumn{1}{r|}{0.97} & 0.60 & 0.60 \\ \cline{3-10}
 && Overall & 0.83 & 0.83 & 0.83 & \multicolumn{1}{r|}{0.83} & \multicolumn{1}{r|}{0.98} & 0.62  & 0.61 \\ \bottomrule
    \end{tabular}
    \caption{Detailed results for our model (BART) trained with the variety of input and output templates (\Cref{tab:train_templates}) and comparative model PAED trained and tested on the PersonaExt-PeaCoK dataset. Results are on the test set.}
    \label{tab:results_persona_extract_detailed}
\end{table*}

\begin{table*}
\resizebox{\textwidth}{!}{\begin{tabular}{lllrrrr|r|rr|r}
\toprule
 &&& \multicolumn{4}{c}{Label} & \multicolumn{1}{c}{Head} & \multicolumn{2}{c}{Tail} & \multicolumn{1}{c}{Triplet} \\
 &&& Precision & Recall & F1 & Acc & Acc & Overlap & Acc & Acc \\
\midrule
\multirow{18}{*}{BART} &
\multirow{6}{*}{Greedy Search} & Characteristic & 0.93 & 0.83 & 0.88 & 0.83 & 0.98 & 0.75 & 0.62 & 0.62 \\
 && Experience & 0.78 & 0.73 & 0.75 & 0.73 & 0.92 & 0.77 & 0.67 & 0.64 \\
 && Goal plan & 0.68 & 0.75 & 0.71 & 0.75 & 0.97 & 0.77 & 0.67 & 0.67 \\
 && Routine habit & 0.73 & 0.88 & 0.80 & 0.88 & 0.98 & 0.83 & 0.75 & 0.74 \\
 && Not & 0.82 & 0.69 & 0.75 & 0.69 & 0.69 & 0.69 & 0.69 & 0.69 \\ \cline{3-11}
 && Overall & 0.85 & 0.84 & 0.84 & 0.84 & 0.96 & 0.78 & 0.67 & 0.66 \\
\cline{2-11}
&\multirow{6}{*}{Beam Search} & Characteristic & \textbf{0.93} & 0.83 & 0.88 & 0.83 & 0.98 & 0.75 & 0.62 & 0.62 \\
 && Experience & 0.78 & 0.74 & 0.76 & 0.74 & 0.92 & 0.78 & 0.69 & 0.65 \\
 && Goal plan & 0.65 & 0.77 & 0.70 & 0.77 & 0.97 & 0.78 & 0.69 & 0.69 \\
 && Routine habit & 0.73 & 0.87 & 0.79 & 0.87 & 0.98 & 0.83 & 0.75 & 0.74 \\
 && Not & 0.83 & 0.68 & 0.75 & 0.68 & 0.68 & 0.68 & 0.68 & 0.68 \\ \cline{3-11}
 && Overall & 0.85 & 0.83 & 0.84 & 0.83 & 0.96 & 0.78 & 0.67 & 0.66 \\
\cline{2-11}
&\multirow{6}{*}{Diverse Beam Search} & Characteristic & 0.93 & 0.83 & 0.88 & 0.83 & 0.98 & 0.76 & 0.63 & 0.62 \\
 && Experience & 0.78 & 0.74 & 0.76 & 0.74 & 0.92 & 0.78 & 0.67 & 0.64 \\
 && Goal plan & 0.67 & 0.76 & 0.71 & 0.76 & 0.98 & 0.77 & 0.67 & 0.67 \\
 && Routine habit & 0.73 & 0.88 & 0.79 & 0.88 & 0.98 & 0.83 & 0.75 & 0.74 \\
 && Not & 0.85 & 0.63 & 0.73 & 0.63 & 0.63 & 0.63 & 0.63 & 0.63 \\ \cline{3-11}
 && Overall & 0.85 & 0.83 & 0.83 & 0.83 & 0.96 & 0.78 & 0.67 & 0.66 \\
\midrule
\multirow{12}{*}{PAED} 
& \multirow{6}{*}{(fine-tuned)} & Characteristic & 0.87 & 0.91 & \textbf{0.89} & 0.91 & 0.97 & 0.77 & 0.61 & 0.61 \\
 && Experience & 0.70 & 0.68 & 0.69 & 0.68 & 0.88 & 0.74 & 0.65 & 0.60 \\
 && Goal plan & 0.73 & 0.75 & 0.74 & 0.75 & 0.97 & 0.75 & 0.67 & 0.67 \\
 && Routine habit & 0.81 & 0.74 & 0.78 & 0.74 & 0.96 & 0.72 & 0.62 & 0.60 \\
 && Not & 0.69 & 0.65 & 0.67 & 0.65 & 0.66 & 0.66 & 0.66 & 0.65 \\ \cline{3-11}
 && Overall & 0.83 & 0.83 & 0.83 & 0.83 & 0.95 & 0.75 & 0.62 & 0.61 \\
\cline{2-11}
& \multirow{6}{*}{(zeroshot)} & Characteristic & 0.87 & 0.01 & 0.01 & 0.01 & 0.02 & 0.01 & 0.01 & 0.00 \\
 && Experience & 0.00 & 0.00 & 0.00 & 0.00 & 0.01 & 0.00 & 0.00 & 0.00 \\
 && Goal plan & 0.48 & 0.42 & 0.45 & 0.42 & 0.42 & 0.40 & 0.39 & 0.39 \\
 && Routine habit & 0.00 & 0.00 & 0.00 & 0.00 & 0.03 & 0.02 & 0.02 & 0.00 \\
 && Not & 0.04 & \textbf{1.00} & 0.08 & \textbf{1.00} & \textbf{1.00} & \textbf{1.00} & \textbf{1.00} & \textbf{1.00} \\ \cline{3-11}
 && Overall & 0.52 & 0.06 & 0.03 & 0.06 & 0.08 & 0.07 & 0.07 & 0.06 \\
\bottomrule
\end{tabular}}
\caption{Detailed per-label results on \pep\xspace test split.}
\label{tab:results_detailed_personaext_peacok}
\end{table*}

\subsection{Comparative Model: PAED}
\label{app:paed_reproduce}
Since we need the model to generate tail phrases, we trained PAED on the modified \pxt dataset with tail phrases \Cref{app:data-ppe}. 
We train two versions of the model: zero-shot on \pxt (with tail phrases) and fine-tuned on \pep.
We use the authors' released codebase \citep{zhu-etal-2023-paed} and their default hyperparameters (we release our copy of their codebase for reproducibility). 
% The model uses 128 sequence length but we use BART's original context window of 1024. ???

\paragraph{Zero-shot}
We trained the model in the zero-shot setting with 10 unseen labels and seed 0 (\Cref{tab:results_paed_personaext_zero}).
The reported results for zero-shot PAED on their \pxt dataset are as follows: 0.40 for $n=5$, 0.32 for $n=10$, and 0.23 for $n=15$.
The triplet accuracy score of $0.34$ is close to the reported score of $0.32$, despite the tail phrase modification of the dataset.

\paragraph{Non-zero shot (fine-tuned)}
To evaluate the PAED model equally with our model, we train PAED in a non-zero-shot setting on \pep.
We use the same default settings as in the released PAED codebase.
% Add PAED details?

\begin{table*}
\centering
\begin{tabular}{lrrrr|r|r|r}
\toprule
 & \multicolumn{4}{c}{Label} & \multicolumn{1}{c}{Head} & \multicolumn{1}{c}{Tail} & Triplet \\
 & Precision & Recall & F1 & Acc. & Acc. & Acc. & Acc. \\
\midrule
dislike & 0.79 & 0.57 & 0.67 & 0.57 & 0.92 & 0.50 & 0.32 \\
favorite\_place & 0.38 & 0.50 & 0.43 & 0.50 & 1.00 & 0.72 & 0.42 \\
get\_along & 0.00 & 0.00 & 0.00 & 0.00 & 0.89 & 0.71 & 0.00 \\
have\_no\_children & 0.00 & 0.00 & 0.00 & 0.00 & 1.00 & 0.79 & 0.00 \\
have\_no\_sibling & 0.00 & 0.00 & 0.00 & 0.00 & 1.00 & 0.90 & 0.00 \\
have\_sibling & 0.51 & 0.89 & 0.65 & 0.89 & 0.85 & 0.64 & 0.56 \\
like\_color & 0.00 & 0.00 & 0.00 & 0.00 & 0.99 & 0.80 & 0.00 \\
like\_general & 0.49 & 0.60 & 0.54 & 0.60 & 0.98 & 0.40 & 0.25 \\
used\_to & 0.18 & 0.16 & 0.17 & 0.16 & 1.00 & 0.68 & 0.11 \\
want\_job & 0.84 & 0.88 & 0.86 & 0.88 & 0.99 & 0.73 & 0.68 \\
\midrule
Overall & 0.52 & 0.58 & 0.54 & 0.58 & 0.95 & 0.58 & 0.34 \\
\bottomrule
\end{tabular}
    \caption{Results for original PAED model on PersonaExt (with ``tail phrases'' (see \Cref{app:data-ppe})) with 10 unseen labels and seed 0. We refer to this model as PAED (zero-shot). }
    \label{tab:results_paed_personaext_zero}
\end{table*}

\begin{table*}
\centering
\begin{tabular}{@{}lllll|ll|l@{}}
\toprule
 & \multicolumn{4}{c}{Relation} & \multicolumn{1}{c}{Head} & \multicolumn{1}{c}{Tail} & \multicolumn{1}{c}{Overall} \\
\multicolumn{1}{c}{} & \multicolumn{1}{c}{Precision} & \multicolumn{1}{c}{Recall} & \multicolumn{1}{c}{F1} & \multicolumn{1}{c}{Acc.} & \multicolumn{1}{c}{Acc.} & \multicolumn{1}{c}{Acc.} & \multicolumn{1}{c}{Acc.} \\
\toprule
Characteristic & 0.92 & 0.79 & 0.85 & 0.79 & 0.99 & 0.80 & 0.65 \\
Experience & 0.33 & 0.90 & 0.48 & 0.90 & 0.90 & 0.83 & 0.75 \\
Goal or Plan & 0.41 & 0.85 & 0.55 & 0.85 & 0.99 & 0.78 & 0.77 \\
Routine or Habit & 0.77 & 0.65 & 0.71 & 0.65 & 0.97 & 0.80 & 0.57 \\
Not & 0.46 & 0.83 & 0.59 & 0.83 & 0.97 & 0.74 & 0.72 \\
\midrule
Overall & 0.81 & 0.75 & 0.77 & 0.75 & 0.98 & 0.80 & 0.63 \\
\bottomrule
\end{tabular}
    \caption{PAED results on the PersonaExt-PeaCoK dataset on the test set without the tail phrase modification.}
    \label{tab:results_paed_ppe_nophrase}
\end{table*}

%%%%%%%%%%%%%%%
\section{Detailed Results}
\label{app:results}
\subsection{\pck-\pxt}
\label{app:results_pck}
The detailed per-label results are in \Cref{tab:results_detailed_personaext_peacok}.

\subsection{LIGHT}
The detailed annotations are in \Cref{tab:results_light_qualitative_annotation}.

\begin{table*}
\resizebox{\textwidth}{!}{\begin{tabular}{lllrrrr|rr|r}
\toprule
&&& \multicolumn{4}{c}{Not Accepted (No)} & \multicolumn{2}{c}{Accepted (Yes)} & Ratio Accepted \\
 &  & & Contradictory & Malformed & Non-specific relation & Unreasonable & Directly & Reasonable \\
\midrule
\multirow[c]{6}{*}{Neutral Removed} & \multirow[c]{3}{*}{NLI} & Beam Search & 8 & 0 & 12 & 10 & 3 & 36 & 0.57 \\
 &  & Diverse Beam Search & 6 & 0 & 8 & 7 & 2 & 31 & 0.61 \\
 &  & Greedy Search & 2 & 0 & 8 & 8 & 1 & 26 & 0.60 \\\cline{2-10}
 & \multirow[c]{3}{*}{Persona-NLI} & Beam Search & 10 & 0 & 51 & 14 & 9 & 75 & 0.53 \\
 &  & Diverse Beam Search & 5 & 3 & 32 & 9 & 8 & 61 & 0.58 \\
 &  & Greedy Search & 3 & 0 & 14 & 6 & 6 & 44 & \textbf{0.68} \\
\midrule
\multirow[c]{4}{*}{Re-ranking} & \multirow[c]{2}{*}{NLI} & Beam Search & 60 & 1 & 85 & 61 & 40 & 320 & 0.63 \\
 &  & Diverse Beam Search & 61 & 4 & 92 & 62 & 38 & 326 & 0.62 \\\cline{2-10}
 & \multirow[c]{2}{*}{Persona-NLI} & Beam Search & 58 & 1 & 99 & 63 & 36 & 313 & 0.61 \\
 &  & Diverse Beam Search & 58 & 4 & 99 & 62 & 37 & 324 & 0.62 \\
 \midrule
 \multirow[c]{3}{*}{Base} & & Beam Search & 57 & 1 & 86 & 61 & 40 & 320 & 0.64 \\
 &  & Diverse Beam Search & 59 & 4 & 92 & 64 & 39 & 326 & 0.62 \\
 &  & Greedy Search & 54 & 4 & 86 & 62 & 44 & 325 & 0.64 \\
\bottomrule
\end{tabular}}
\caption{Detailed results for human evaluation on personas extracted from LIGHT dataset. IAA is 0.85.}
\label{tab:results_light_qualitative_annotation}
\end{table*}

%\begin{table*}
%\resizebox{\textwidth}{!}{\begin{tabular}{l|rrrrr|rr|r}
%\toprule
%& Disputed & \multicolumn{4}{c|}{Not Accepted (No)} & \multicolumn{2}{c|}{Accepted (Yes)} & Ratio Accepted \\
%Character & & Contradictory & Malformed & Non-specific relation & Unreasonable & Directly & Reasonable &  \\
%a masked torturer\_63 & 24 & 37 & 0 & 89 & 3 & 31 & 90 & 0.44 \\
%a messenger\_65 & 76 & 53 & 0 & 168 & 40 & 17 & 395 & 0.55 \\
%bird\_244 & 9 & 13 & 0 & 25 & 9 & 16 & 78 & 0.63 \\
%guard\_603 & 73 & 46 & 1 & 45 & 67 & 18 & 214 & 0.50 \\
%lost traveler\_742 & 8 & 0 & 1 & 51 & 18 & 0 & 346 & 0.82 \\
%monk\_811 & 6 & 0 & 0 & 20 & 0 & 20 & 21 & 0.61 \\
%person\_927 & 77 & 45 & 15 & 163 & 97 & 58 & 314 & 0.48 \\
%pet goldfish\_940 & 9 & 81 & 0 & 58 & 58 & 47 & 112 & 0.44 \\
%pirate\_948 & 224 & 189 & 13 & 242 & 158 & 107 & 1109 & 0.60 \\
%\bottomrule
%\end{tabular}}
%\caption{Per-character annotations on LIGHT dataset.}
%\label{tab:results_light_qualitative_character}
%\end{table*}

\end{document}